%% file: main-5965-Kobayashi.tex
\pdfoutput=1
\documentclass[11pt,a4paper]{article}
\usepackage{times,latexsym}
\usepackage{url}
\usepackage[T1]{fontenc}
\usepackage{graphicx}
\usepackage{multirow}
\usepackage{colortbl}
\usepackage{subcaption}
\usepackage{caption}
\usepackage{fullpage}
\usepackage{tikz}
\usepackage[xcolor,framemethod=tikz]{mdframed}
\usepackage{color}

\newcommand{\seeda}{{\sc{SEEDA}}\xspace}
\newcommand{\mtwo}{$M^{2}$\xspace}
\newcommand{\sentmtwo}{Sent$M^{2}$\xspace}
\newcommand{\ptmtwo}{PT-$M^{2}$\xspace}

\newcommand{\bgfb}[1]{\tikz[baseline=(X.base)]{\node(X)[rectangle, fill=red!60!blue!18, rounded corners, text height=.8ex,text depth=-0.5ex]{\textit{#1}};}}
\newcommand{\bgfc}[1]{\tikz[baseline=(X.base)]{\node(X)[rectangle, fill=green!60!blue!18, rounded corners, text height=.8ex,text depth=-0.5ex]{\textit{#1}};}}

\usepackage[acceptedWithA]{tacl2021v1}

\usepackage{xspace,mfirstuc,tabulary}

\makeatletter
\newcommand{\figcaption}[1]{\def\@captype{figure}\caption{#1}}
\newcommand{\tblcaption}[1]{\def\@captype{table}\caption{#1}}
\makeatother

\newif\iftaclinstructions
\taclinstructionsfalse 
\iftaclinstructions
\renewcommand{\confidential}{}
\renewcommand{\anonsubtext}{(No author info supplied here, for consistency with
TACL-submission anonymization requirements)}
\newcommand{\instr}
\fi

\iftaclpubformat 

\else

\fi

\title{Revisiting Meta-evaluation for Grammatical Error Correction}

\author{
  Masamune Kobayashi$^\diamond$
  \ \ \
  Masato Mita$^\dagger$$^\diamond$
  \ \ \
  Mamoru Komachi$^\ddagger$
  \\
  $^\diamond$Tokyo Metropolitan University, Japan
  \ \ \
  $^\dagger$CyberAgent Inc.
  \ \ \
  $^\ddagger$Hitotsubashi University, Japan
  \\
  \texttt{kobayashi-masamune@ed.tmu.ac.jp,}
  \ \ \
  \texttt{mita\_masato@cyberagent.co.jp,}
  \\
  \texttt{mamoru.komachi@r.hit-u.ac.jp}
}

\date{}

\begin{document}
\maketitle
\begin{abstract}
Metrics are the foundation for automatic evaluation in grammatical error correction (GEC), with their evaluation of the metrics (meta-evaluation) relying on their correlation with human judgments.
However, conventional meta-evaluations in English GEC encounter several challenges including biases caused by inconsistencies in evaluation granularity, and an outdated setup using classical systems. 
These problems can lead to misinterpretation of metrics and potentially hinder the applicability of GEC techniques.
To address these issues, this paper proposes~\seeda, a new dataset for GEC meta-evaluation.
\seeda consists of corrections with human ratings along two different granularities:~\textit{edit-based} and \textit{sentence-based}, covering 12 state-of-the-art systems including large language models (LLMs), and two human corrections with different focuses.
The results of improved correlations by aligning the granularity in the sentence-level meta-evaluation suggest that edit-based metrics may have been underestimated in existing studies.
Furthermore, correlations of most metrics decrease when changing from classical to neural systems, indicating that traditional metrics are relatively poor at evaluating fluently corrected sentences with many edits.
\end{abstract}

\section{Introduction}
Grammatical error correction (GEC) is the task of automatically detecting and correcting errors, including grammatical, orthographic, and semantic errors, within a given sentence. 
The prevailing approach in GEC involves the use of a sequence-to-sequence method~\cite{bryant2023grammatical}.

Automatic evaluation metrics play an important role in the progress of GEC.
These metrics are essential for a fast and efficient improvement cycle of system development because they can replace costly and time-consuming human evaluations and immediately reflect system performance.
GEC has made progress by enabling a fair comparison of performance on a common benchmark using these metrics in shared tasks~\cite{dale-kilgarriff-2011-helping,dale-etal-2012-hoo,ng-etal-2013-conll,ng-etal-2014-conll,bryant-etal-2019-bea}.

GEC metrics are categorized into \textit{edit-based} and \textit{sentence-based} types according to their evaluation granularity, and each has its objectives.
Edit-Based Metrics (EBMs), such as \mtwo~\cite{dahlmeier-ng-2012-better} and ERRANT~\cite{bryant-etal-2017-automatic}, focus on evaluating the quality of the edit itself, whereas Sentence-Based Metrics (SBMs), such as GLEU~\cite{napoles-etal-2015-ground}, evaluate the quality of the entire sentence after correction.
Since the system output consists only of sentences without explicit edits, EBMs require the edit extraction from the system output using any method.
In addition, these metrics are primarily evaluated based on the correlation with human judgment (i.e., \textit{meta-evaluation}).

\input{tables/tab_example_granularity}

Most of the previous meta-evaluations in English GEC have relied on \citet{grundkiewicz-etal-2015-human}'s dataset with human judgments (henceforth, this dataset is referred to as \textit{GJG15}).
However, existing meta-evaluations based on GJG15~\cite{grundkiewicz-etal-2015-human,chollampatt-ng-2018-reassessment,yoshimura-etal-2020-reference,gong-etal-2022-revisiting} have several significant issues.
First, the performance of EBMs may be underestimated due to biases resulting from inconsistencies in evaluation granularity.
As an example of biases, while EBMs assign the lowest score (or the highest score in the sentence-level evaluation) to the uncorrected sentence, sentence-based human evaluation, such as GJG15, assigns scores across the entire range.
Furthermore, according to the actual data in Table~\ref{tab:granularity diff}, since human evaluations may yield different results based on granularity, the GEC evaluation suggests a need to separate evaluations for edits and sentences.
Second, GJG15 is manually evaluated against the set of classical systems in CoNLL-2014 shared task, such as statistical machine translation approach~\cite{junczys-dowmunt-grundkiewicz-2014-amu}, and classifier-based approach~\cite{rozovskaya-etal-2014-illinois}.
Therefore, the gap between the classical systems in GJG15 and the current modern GEC systems based on deep neural networks limits the applicability of meta-evaluation.
Third, a single correlation from the current fixed set of systems may not sufficiently capture the performance of metrics, leading to the possibility of drawing incorrect conclusions.
For example, \citet{deutsch-etal-2021-statistical}'s study on meta-evaluation of summarization revealed that certain metrics can exhibit a spectrum of correlation values, ranging from weak negative to strong positive correlations.
\citet{mathur-etal-2020-tangled}'s study also showed that outlier systems have a strong influence on correlations in a meta-evaluation of machine translation.
Therefore, we are concerned that a similar scenario could occur in the GEC.

To address these issues, we propose \seeda,~\footnote{\seeda stands for \textbf{S}entence-based and \textbf{E}dit-based human \textbf{E}valuation \textbf{DA}taset for GEC. We have made this dataset publicly available at \url{https://github.com/tmu-nlp/SEEDA}.} a new dataset to improve the validity of meta-evaluation in English GEC.
Specifically, we carefully designed \seeda to address the first and second issues by performing human evaluations corresponding to two different granularity metrics (i.e., EBMs and SBMs), covering 12 state-of-the-art system corrections including large language models (LLMs), and two human corrections with different focuses (\S \ref{section3} and \S \ref{section4}).
Also, through meta-evaluation using \seeda, we investigate whether EBMs, such as \mtwo and ERRANT, are underestimated and demonstrate how the correlation varies between classical systems and neural systems (\S \ref{section6}).
Furthermore, to address the third issue, we investigate the inadequacy of GEC meta-evaluation based solely on a single correlation by analyzing the presence of outliers and using window analysis (\S \ref{section7}).
Finally, we discuss best practices and provide recommendations for future researchers to properly meta-evaluate GEC metrics and evaluate their GEC models (\S \ref{section8}).

Our contributions are summarized as follows.
(1) We construct a new dataset that allows for bias-free meta-evaluation that fits modern neural systems.
(2) The dataset analysis shows variations in sentence-level human evaluation results depending on the evaluation granularity.
(3) We identified two findings through meta-evaluation: aligning the granularity between human evaluation and metric enhances correlations, and correlations for classical and neural systems are different.
(4) Investigating the influence of outliers and system sets, we discovered that a meta-evaluation of a single setting cannot analyze the detailed characteristics of the metric.
We also found that existing metrics lack the precision to differentiate between the performances of top-tier systems.

\section{Related work}
\paragraph{Meta-evaluation:}
\citet{grundkiewicz-etal-2015-human} proposed a dataset (GJG15) with sentence-based human ratings for system outputs in the CoNLL-2014 test set and found that \mtwo has a moderate positive correlation with human judgments.
Simultaneously, \citet{napoles-etal-2015-ground} constructed a dataset by performing a similar human evaluation and observed that their proposed metric, GLEU, has a stronger correlation than \mtwo.
Both studies found no correlation with I-measure~\cite{felice-briscoe-2015-towards}.
\citet{chollampatt-ng-2018-reassessment} carried out significance tests between various metrics using GJG15.
They concluded that there was no clear distinction in performance between \mtwo and GLEU, with I-measure proving to be the most robust metric.
However, these experiments are based on classical systems and thus deviate from modern neural systems.
MAEGE proposed by \citet{choshen-abend-2018-automatic} applies multiple partial edits to the uncorrected sentence and assigns pseudo-scores based on the number of edits, aiming for a meta-evaluation independent of human evaluation. 
MAEGE does not consider system outputs and human evaluations, so it should be distinguished from existing meta-evaluations that rely on humans.
Moreover, since it does not account for errors that machines might make but humans wouldn't, the need for human evaluation against outputs persists.
Furthermore, \citet{napoles-etal-2019-enabling} constructed GMEG-Data by performing human judgments using continuous scales on the CoNLL-2014 test set and three domain-specific datasets.
Their findings highlighted diverse correlations across the different domains.
They explored neural systems, but these deviate from mainstream systems pretrained with pseudo data and fine-tuned based on the transformer~\cite{vaswani2017attention}.
While \seeda offers greater validity due to its focus on contemporary target systems and the evaluation granularity, GMEG-Data has the advantage of allowing meta-evaluation using the entire CoNLL-2014 benchmark in various domains.

\paragraph{Reference-based evaluation:}
In the evaluation of GEC, commonly used metrics rely on reference sentences.
Some of the most prevalent metrics include \mtwo, ERRANT, and GLEU.
Both \mtwo and ERRANT calculate $F_{0.5}$ score by comparing the edits in the corrected sentence to those in the reference.
In contrast, GLEU assesses based on the matching of N-grams between the corrected sentence and reference.
I-measure evaluates the degree of improvement from the original sentence using the weighted precision of edits.
There are also newer metrics like GoToScorer~\cite{gotou-etal-2020-taking}, which takes into account the difficulty of corrections, and \ptmtwo~\cite{gong-etal-2022-revisiting}, which extends \mtwo (and ERRANT) with pretraining-based metrics.
It is worth noting that these reference-based evaluations can lose validity with limited reference coverage.

\paragraph{Reference-less evaluation:}
Evaluations without reference sentences aim to overcome the coverage issues.
GBM~\cite{napoles-etal-2016-theres} estimates grammaticality by identifying the number of errors in a sentence.
However, it may be less sensitive to semantic changes.
To address this limitation, GFM~\cite{asano-etal-2017-reference} was proposed.
It incorporates sub-metrics to estimate grammaticality, fluency, and meaning preservation.
Additionally, USim~\cite{choshen-abend-2018-reference} was developed to specifically estimate semantic faithfulness.
SOME~\cite{yoshimura-etal-2020-reference} draws inspiration from GFM and optimizes each sub-metric based on human evaluation using BERT~\cite{devlin-etal-2019-bert}.
Scribendi Score~\cite{islam-magnani-2021-end} relies on various factors, including GPT-2 perplexity, token sort ratio, and Levenshtein distance ratio, to evaluate correction quality.
IMPARA~\cite{maeda-etal-2022-impara} fine-tunes BERT using only parallel data to quantify the impact of corrections.
In terms of quality estimation, \citet{chollampatt-ng-2018-neural} introduced the first neural approach that does not rely on handcrafted features, while \citet{liu-etal-2021-neural} considered interactions between hypotheses using inference graphs.

\input{figures/fig_m2_wer}

\section{The \seeda dataset}\label{section3}
The \seeda dataset consists of corrections annotated with human ratings along two different evaluation granularities: edit- and sentence-based, covering 12 state-of-the-art neural systems including LLMs, and two human corrections.
The \seeda dataset is denoted as \seeda-E for edit-based evaluation and \seeda-S for sentence-based evaluation.
In this section, we describe the \seeda dataset, how we generated the corrections (\S\ref{section3.1}), and how we collected the annotations (\S\ref{section3.2}).
We use the CoNLL-2014 test set~\cite{ng-etal-2014-conll} as our input data, consisting of test essays and their error annotations.
The test essays are written by non-native English-speaking students from the National University of Singapore and cover two genres: genetic testing and social media. 
Error annotations for the test essays are conducted by two native English speakers.
The data comprises a total of 50 essays, consisting of 1,312 sentences and 30,144 tokens.

\subsection{GEC Systems}\label{section3.1}
To align with the current setting in GEC, we collect corrections using two mainstream neural-based approaches: \textit{sequence-to-sequence} and \textit{sequence tagging}~\cite{bryant2023grammatical}.
To investigate how highly discriminating current metrics are, top-tier systems should be included among the target systems.
This includes the LLMs which have received increased attention in recent years.
Following these requirements, we carefully selected 11 systems, ensuring that the count is no less than the number of systems in GJG15.
Among these, eight systems are sequence-to-sequence models that generate each token autoregressively, TemplateGEC~\cite{li-etal-2023-templategec}, TransGEC~\cite{fang-etal-2023-transgec}, T5~\cite{rothe-etal-2021-simple}, LM-Critic~\cite{yasunaga-etal-2021-lm}, BART~\cite{DBLP:journals/corr/abs-1910-13461}, BERT-fuse~\cite{kaneko-etal-2020-encoder}, Riken Tohoku~\cite{kiyono-etal-2019-empirical}, and UEDIN-MS~\cite{grundkiewicz-etal-2019-neural}.
The remaining three systems are sequence tagging models that predict edit tags in parallel, GECToR-ens~\cite{tarnavskyi-etal-2022-ensembling}, GECToR-BERT~\cite{omelianchuk-etal-2020-gector}, and PIE~\cite{awasthi-etal-2019-parallel}.
Following the recent LLMs trend, we consider GPT-3.5 (\texttt{text-davinci-003}) for two-shot learning~\cite{coyne2023analyzing}.
We included INPUT (source from the CoNLL-2014 test set) since GEC evaluation requires consideration of uncorrected sentences.
We also consider REF-M (minimal edit references by experts) and REF-F (fluency edit references by experts), which are introduced by \citet{sakaguchi-etal-2016-reassessing}, to compare the system performance with human correction, bringing to the total to 15 sentence sets.

\input{figures/fig_overview}

Figure~\ref{fig:system analysis} shows the \mtwo Score ($F_{0.5}$)~\footnote{In GEC, it is common to use $F_{0.5}$, where Precision is given twice the importance of Recall~\cite{ng-etal-2014-conll,bryant-etal-2019-bea}. This is because, in the practical usage of GEC systems, not correcting is not as detrimental as making incorrect corrections. Additionally, in the context of language acquisition where minimizing incorrect feedback is desirable, this weighting is reasonable~\cite{nagata:coling2010}.} and word edit rate for classical systems in GJG15, neural systems in \seeda, and human sentences.
Comparing these systems, neural systems in \seeda show a higher number of edits and demonstrate better correction performance from the perspective of \mtwo.
This performance comparison utilizes the most common GEC evaluation method, reproducing results reported in existing studies.
On the other hand, this performance comparison contains intuitive contradictions, such as the lower performance of human-corrected sentences and LLMs.
Therefore, we investigate and report how the modern system comparison deviates from human judgments (\S \ref{section4.2}).
Note that few-shot learning such as GPT-3.5 is known to be not grounding to target sentences as compared to finetuned models and may produce fluent but lengthy correction sentences that do not preserve the meaning of the source~\cite{maynez-etal-2023-benchmarking}.

\subsection{Annotation scheme}\label{section3.2}
\paragraph{Edit-based human evaluation:}
In the edit-based human evaluation, we evaluate only for edits in the system output.
We perform a step-by-step sequence labeling using the doccano annotation tool~\cite{doccano}.
In the edit-based human evaluation, we decided to divide the process into two steps to avoid complicating the annotation process.

Figure~\ref{fig:overview} shows an overview of the annotation flow and an example of edit-based human evaluation.
In Step 1, the detection of errors in the source and checking for edits in the output are performed.
During the initial error detection, annotators refer to 25 error categories by \citet{bryant-etal-2017-automatic} to identify error locations in the source, enabling them to label errors at the minimal unit level.
In the subsequent Edit checking, annotators perform a binary decision to determine whether they would like to apply the edits in the output to improve the source or not.
To reduce annotation costs, ERRANT is used for extracting edits.
When there are conflicting edits (e.g., subject-verb agreement error), the one that aligns with the context is deemed effective, while the other is considered ineffective.
Furthermore, for edits that depend on each other (e.g., [law's→ ] and [ →of the law] in Figure~\ref{fig:overview}), each is assigned an independent label, but they are deemed effective only if all dependent edits are present.
In Step 2, the annotator performs a binary decision to determine whether each edit in the output effectively corrects the errors found in Step 1.
Finally, we compute $F_{0.5}$ based on Precision and Recall~\footnote{Note that Precision and Recall are computed at different levels of granularity.} for each corrected sentence and subsequently rank the set of corrected sentences accordingly.
The supplementary information about the annotation is provided in Appendix \ref{appendix:a}.

\input{tables/tab_human_rankings}
\input{tables/tab_stats}

\paragraph{Sentence-based human evaluation:}
Following \citet{grundkiewicz-etal-2015-human}, sentence-based human evaluation is performed using the Appraise evaluation scheme~\cite{federmann-2010-appraise}.
Annotators read the context in the same way as edit-based human evaluation.
And then, the corrected sentences are relatively ranked, allowing the same rank from the best to the worst. 
The judgment of whether a sentence is good or bad is left to the subjectivity of each annotator.

\paragraph{Annotator and sampling method:}
Each annotation was performed by three native English speakers with extensive knowledge of the language.
To observe differences by evaluation granularity, they are responsible for the same set of edit-based and sentence-based annotations.
Following \citet{grundkiewicz-etal-2015-human}, we sample 200 subsets from the 1312 correction sets against the CoNLL-2014 test set using a parameterized distribution that favors more diverse outputs.
To measure inter- and intra-annotator agreements, we duplicated at least 12.5\% of the subset.
One subset may contain up to five sentences, and the annotator creates a ranking from those sentences.

\section{Dataset analysis}\label{section4}
In this section, we analyze \seeda with a focus on evaluation granularity.
First, we present the dataset statistics (\S \ref{section4.1}).
Second, we produce human rankings for the system using rating algorithms to conduct system-level meta-evaluation (\S \ref{section4.2}).
Third, we quantitatively analyze to discern any disparities in human evaluations across different evaluation granularities (\S \ref{section4.3}).

\subsection{Dataset statistics}\label{section4.1}
Table~\ref{tab:dataset statistics} presents the statistics for pairwise judgments by annotators.
Each annotator has created 200 rankings for each subset, resulting in a total of 600 rankings.
We take all combinations of all two sentences (A, B) for ranking, make a pairwise judgment (A$>$B, A$=$B, A$<$B), and count their numbers.
To investigate the frequency of duplicate corrections, the raw data was expanded by treating systems that produced the same output independently.
As a result, the number of pairwise evaluations increased significantly.
This finding, similar to classical systems in \citet{grundkiewicz-etal-2015-human}, suggests that even high-performing neural systems that make many edits often generate duplicated corrections.
Moving forward, experiments will be conducted using raw data of pairwise judgments.
Table~\ref{tab:annotator agreement} shows average inter- and intra-annotator agreements.
Cohen's kappa coefficient ($\kappa$)~\cite{cohen_coefficient_1960} is used to measure the agreement.
In comparison to the results in \citet{grundkiewicz-etal-2015-human}, the high inter- and intra-annotator agreement indicates that the annotators were able to provide more consistent evaluations.

\input{tables/tab_IAA}
\input{figures/fig_scatterplots}

\subsection{Human rankings}\label{section4.2}
Following \citet{grundkiewicz-etal-2015-human}, we employed two rating algorithms, TrueSkill (TS) from \citet{sakaguchi-etal-2014-efficient} and Expected Wins (EW) from \citet{bojar-etal-2013-findings}, to create human rankings based on pairwise judgments.
Table~\ref{tab:system ranking} shows the human rankings generated using TS for both edit-based and sentence-based evaluations. 
In contrast to classical systems in GJG15, all the neural systems receive ranks surpassing INPUT.
This indicates a tendency of these systems to improve uncorrected sentences through correction.
Systems based on GPT and T5 architectures (e.g., GPT-3.5, T5, TransGEC) achieve higher rankings than REF-M.
This suggests the potential of these systems to offer corrections that might even surpass human capabilities.

\subsection{Difference in human evaluation by granularity}\label{section4.3}
We perform a quantitative analysis of the variations in human evaluation based on granularity.
To measure sentence-level agreement, we calculate the average intra-annotator $\kappa$ between edit-based and sentence-based evaluations.
The result, a modest 0.36, indicates low agreement.
On the other hand, the system-level $\kappa$ using pairwise judgments from the human rankings stands at a much higher 0.83, revealing negligible disparity.
This indicates a pronounced difference in sentence-level evaluation, but a relatively minor one in system-level evaluation.
This suggests that biases are more prominent at the sentence-level meta-evaluation.

\input{tables/tab_sys-level_meta-eval}
\input{tables/tab_effect_outlier}

\section{Baseline metrics}\label{section5}
We target 11 GEC metrics for meta-evaluation, including EBMs (\S~\ref{section5.1}) and SBMs (\S~\ref{section5.2}).

\subsection{Edit-based metrics}\label{section5.1}
\paragraph{\mtwo \citep{dahlmeier-ng-2012-better}.} It compares the edits in the corrected sentence with those in the reference. It dynamically searches for edits to optimize alignment with the reference edits using Levenshtein alignment~\cite{Levenshtein1966bcc}.
\paragraph{\sentmtwo.} It is a variant of \mtwo that calculates $F_{0.5}$ score at the sentence level.
\paragraph{\ptmtwo \citep{gong-etal-2022-revisiting}.} It is a hybrid metric that combines \mtwo and BERTScore~\cite{DBLP:journals/corr/abs-1904-09675}.
It can measure the semantic similarity between pairs of sentences, not just comparing edits.
\paragraph{ERRANT \citep{bryant-etal-2017-automatic}.} It is similar to \mtwo but differs in that it uses linguistically enhanced Damerau-Levenshtein alignment for extracting edits.
It is characterized by its ability to calculate $F_{0.5}$ score for each error type.
\paragraph{SentERRANT.} It is a variant of ERRANT that computes sentence-level $F_{0.5}$ score.
\paragraph{PT-ERRANT.} It is a variant of \ptmtwo where the base metric has been changed from \mtwo to ERRANT.
\paragraph{GoToScorer \citep{gotou-etal-2020-taking}.} It calculates $F_{0.5}$ score while considering the difficulty of correction.
The difficulty is calculated based on the number of systems that were able to correct the error.

\subsection{Sentence-based metrics}\label{section5.2}
\paragraph{GLEU \citep{napoles-etal-2015-ground}.} 
It is based on the commonly used BLEU~\cite{papineni-etal-2002-bleu} in machine translation.
It rewards N-grams in the output that match the reference but are not in the source while penalizing N-grams in the source that do not match the reference.
For better evaluations, we use GLEU without tuning~\cite{DBLP:journals/corr/NapolesSPT16}.
\paragraph{Scribendi Score \citep{islam-magnani-2021-end}.} It evaluates by combining the perplexity calculated by GPT-2~\cite{radford_language_2019}, token sort ratio and Levenshtein distance ratio.
\paragraph{SOME \citep{yoshimura-etal-2020-reference}.} It optimizes human evaluations by fine-tuning BERT separately for each of the following criteria: grammaticality, fluency, and meaning preservation.
\paragraph{IMPARA \citep{maeda-etal-2022-impara}.} It combines a quality estimation model fine-tuned with parallel data using BERT and a similarity model to consider the impact of edits.

\section{Revisiting meta-evaluation for GEC}\label{section6}
We investigate how correlations are affected by resolving granularity inconsistencies and are changed from classical systems to modern neural systems through system-level (\S \ref{section6.1}) and sentence-level (\S \ref{section6.2}) meta-evaluations.
Figure \ref{fig:scatterplot} shows the scatter plots of the human evaluation and the metric scores, indicating that uncorrected sentences (INPUT) and fluently corrected sentences (REF-F, GPT-3.5) stand out as outliers and influence the correlation.
Therefore, we consider 12 systems, deliberately excluding uncorrected sentences (INPUT) and sentences with fluently corrected sentences (REF-F, GPT-3.5).
We calculate metric scores on the subset targeted in human evaluations.

\subsection{System-level meta-evaluation}\label{section6.1}
\paragraph{Setup:}
For our system-level meta-evaluation, we report correlation using system scores obtained from human rankings.
Metrics such as \mtwo, \ptmtwo, ERRANT, GoToScorer, and GLEU can calculate system scores, while other metrics use the average of sentence-level scores as the system score.
We use Pearson correlation (r) and Spearman rank correlation ($\rho$) to measure the closeness between the metric and human evaluation.

\paragraph{Result:}
According to the system-level meta-evaluation results in Table~\ref{tab:base meta-evaluation}, it is evident that aligning the granularity between the metrics and human evaluation improves the correlation for EBMs in \seeda-E, while the correlation for SBMs in \seeda-S tends to decrease.
One reason for the inconsistent results even when the granularity is aligned is that system-level human evaluations exhibit relatively small variations across different evaluation granularities.

We discovered that as we move from classical systems to neural systems, correlations for all metrics---except GoToScorer and GLEU---decrease through a comparison between GJG15 and \seeda-S.
This result suggests that the majority of current metrics cannot adequately evaluate the more extensively edited and fluent corrections produced by neural systems, in contrast to those generated by classical systems.
In the meta-evaluation results of GJG15, comparing it with existing studies~\cite{grundkiewicz-etal-2015-human,choshen-abend-2018-automatic} is unfeasible, as the exclusion of INPUT has been implemented to alleviate scoring bias between EBMs and sentence-based human evaluation.

\input{figures/fig_window_analysis}

\subsection{Sentence-level meta-evaluation}\label{section6.2}
\paragraph{Setup:}
In sentence-level meta-evaluation, we use pairwise judgments in Table~\ref{tab:dataset statistics} to calculate correlations.
We use Kendall's rank correlation ($\tau$) and Accuracy (Acc) to measure the performance of the metrics.
Kendall ($\tau$) can measure performance in the common use case of comparing corrected sentences to each other.

\paragraph{Result:}
In contrast to the system-level results, sentence-level meta-evaluations showed more significant improvements in correlations when the granularity was aligned.
The substantial variation in sentence-level human evaluations based on granularity likely contributed to more consistent results.
In other words, it became evident that correlations in sentence-level meta-evaluation are underestimated when granularity is not aligned.

When we compared GJG15 and \seeda-S, we observed a decrease in correlations for most metrics, especially in EBMs, similar to the system-level results.
Consistently high correlations were found for SOME and IMPARA, indicating the effectiveness of fine-tuned BERT.

\section{Further analysis}\label{section7}
As further analysis, we investigate the influence of outliers (\S\ref{section7.1}) and variations in the system set (\S\ref{section7.2}) on the correlation of the metric.
We test the hypothesis on which this study focuses, that there may be a range of correlations in flexible settings in GEC.
Based on the best practices obtained in \S\ref{section6}, granularity will be aligned in subsequent meta-evaluations.

\subsection{Influence of outliers}\label{section7.1}
Table \ref{tab:outlier meta-evaluation} shows the results when the uncorrected sentences (INPUT) and/or fluently corrected sentences (REF-F, GPT-3.5) are added to the base meta-evaluation excluding outliers (\S\ref{section6}).

\paragraph{System-level analysis:}
The system-level results show that simply considering INPUT increases the correlations for most metrics to the point where comparisons are difficult.
This suggests that INPUT serves as a strong outlier that skews the correlation positively and prevents accurate meta-evaluation.
One of the reasons is that most EBMs assign the lowest score to INPUT, which also ranks the lowest in human evaluations.
Therefore, in the meta-evaluation using neural models, it was demonstrated that a fair comparison cannot be made when considering the INPUT.

On the other hand, the addition of REF-F and GPT-3.5 shows a sharp drop in overall correlation.
The results suggest that metrics other than SOME and IMPARA cannot properly assess fluently corrected sentences.
Increasing references to commonly used metrics (\mtwo, ERRANT, GLEU) improves the correlation slightly, but still does not provide the same evaluation as humans.
The same tendency as in the \citet{maynez-etal-2023-benchmarking}'s study was observed that the overlap-based metric does not correctly evaluate LLMs for few-shot learning.

\paragraph{Sentence-level analysis:}
The results in the sentence-level meta-evaluation showed a similar trend as system-level results but with some differences.
Adding INPUT improved correlations for most metrics, but both GoToScorer and Scribendi Score have decreased, which may be attributed to the inability to properly perform sentence-based evaluation.
Furthermore, when adding REF-F and GPT-3.5, not only did many metrics show a decrease in correlation, but SOME and IMPARA also exhibited a slight reduction in correlation.

The improved correlations in \mtwo (+Min) and GLEU (+Min), when REF-F and GPT-3.5 were added, indicate that the fluency correction may no longer be an outlier for commonly used metrics if the low coverage of reference-based evaluation is mitigated.
To address the issue of reference coverage, an approach similar to \citet{choshen-abend-2018-automatic}, which involves splitting and combining edits for each reference, could potentially enhance the effective utilization of references.
However, the result that fluency edit references were useful only for GLEU suggests that fluent edit references may be effective on an N-gram basis, but not on an edit extraction basis.
As one of the reasons, we can consider the difficulties and complexities in edit extraction for fluent sentences in EBMs, as well as the inability to address the low coverage of three fluent references.

\subsection{Influence of variations in the system set}\label{section7.2}
Next, we investigate the extent to which the correlation of the metrics varies with changes in a system set.
To create a difficult setting for the metric, correlations are computed for a set of systems with close performance by sorting the systems in order of human ranking.
Figure~\ref{fig:window analysis} shows the variation in correlations using window analysis.
What is common for most metrics is that Pearson (r) tends to be highly variable from positive to negative for evaluation of four systems, but relatively stable for evaluation of eight systems.
This suggests that most metrics do not have enough precision to identify performance differences in a set of high-performance neural systems.
Therefore, there is still a need to develop better metrics that allow precise evaluation.
Furthermore, \mtwo, ERRANT, and GLEU were often uncorrelated or negatively correlated, indicating that the commonly used metrics do not have high robustness.
On the other hand, the BERT-based metrics were found to maintain relatively high correlations, with SOME in particular being the most robust.
Kendall ($\tau$) has a large number of samples for pairwise judgments, so there is no significant change.

\section{Discussion}\label{section8}
We provide a more practical guideline for meta-evaluation (\S \ref{section8.1}) and evaluation (\S \ref{section8.2}) methodologies in future GEC research by considering the experimental results so far.

\subsection{Towards valid meta-evaluation in GEC}\label{section8.1}
We recommend that meta-evaluation be conducted at each evaluation granularity in GEC.
Specifically, EBMs should use SEEDA-E, and SBMs should use SEEDA-S.
The meta-evaluation using \seeda should use the 12 systems as a baseline, excluding outliers, and add REF-F and GPT-3.5 if you want to find out how well the metrics can evaluate fluent corrections.
This allows meta-evaluation for the modern neural system without the bias of the granularity.
Additionally, conducting experiments with various methodologies is crucial to validate the characteristics of metrics. 
Therefore, experiments using GMEG-Data for domain-specific meta-evaluation of SBMs and meta-evaluation by MAEGE, irrespective of granularity, should be considered if resources permit.

The further analysis in \S \ref{section7}, which yielded results unavailable in \S \ref{section6}, demonstrates that conducting meta-evaluation for only a single setting is inadequate in GEC.
Therefore, it is necessary to measure correlations across multiple experimental settings, considering the presence of outliers and more realistic sets of systems with similar performance. 
Additionally, achieving meta-evaluation reliability in GEC using confidence intervals for correlations, like \citet{deutsch-etal-2021-statistical}'s study, is considered important.
Furthermore, annotation based on Multidimensional Quality Metrics~\cite{Lommel:2014} can take into account error types and severity, potentially providing interesting insights when compared to results from WMT~\cite{freitag:wmt2021,freitag:wmt2022}.

\subsection{Best practices for GEC evaluation}\label{section8.2}
We recommend the use of both EBMs and SBMs in GEC.
In light of the trend toward more fluent correcting systems such as the GPT model, the current combination of the CoNLL-2014 test set and \mtwo will no longer be adequate for proper evaluation.
Therefore, it is essential to use high correlation metrics, such as SOME or IMPARA, in addition to \mtwo, to enable the evaluation of LLMs and achieve a more human-like and robust evaluation.
Alternatively, exhaustive fluency references should be prepared to improve \mtwo correlations, or datasets such as JFLEG~\cite{napoles-etal-2017-jfleg} that can account for fluency should be used.
Furthermore, using LLMs, as reported in recent studies~\cite{chiang-lee-2023-large,Liu2023benchmarking,kocmi:eamt2023} as an effective evaluator for other generative tasks, may also prove beneficial in GEC.
If resources allow, it would be good to conduct additional human evaluations.

EBMs and SBMs each have different strengths.
EBMs can calculate Precision, Recall, and F-score, allowing a detailed evaluation of the system performance.
In terms of second language acquisition, the evaluation of each edit provides information about the error location, type, and amount, which can improve the quality of feedback and learning efficiency.
Most SBMs, on the other hand, can evaluate without references, circumventing the problem of underestimating corrections that are limited by the coverage of references.
Also, unlike EBMs, SBMs do not automatically give the lowest score to uncorrected sentences.
This allows for a quantifiable measurement to determine whether a sentence has been improved or worsened as a result of correction.

\input{figures/fig_screenshot}

\section{Conclusion}
To address issues in conventional meta-evaluation in English GEC, we construct a meta-evaluation dataset (\seeda) consisting of corrections with human ratings along two different evaluation granularities, covering 12 state-of-the-art system corrections including LLMs, and two human corrections with different focuses.
The dataset analysis reveals that the results of sentence-level human evaluation differ between granularities and that GEC systems based on GPT and T5 can correct as well as or better than humans.
Also, through meta-evaluation using \seeda, we demonstrate that EBMs may be underestimated in existing meta-evaluations and that matching the evaluation granularity of metrics with human evaluations tends to improve sentence-level correlations.
By further analysis, we discovered the uncertainty of conclusions based on a single correlation and found that most metrics lacked the precision to distinguish differences among high-performance neural systems.
Finally, we propose a methodology for meta-evaluation and evaluation in GEC.
We hope that this paper contributes to further advancements in GEC.

\appendix
\section{Supplement of annotations}\label{appendix:a}
Figure~\ref{fig:screenshot} shows a screenshot of doccano used in the edit-based human evaluation.
The source is enclosed in a <t> tag, and each corrected sentence is emphasized with a <s> tag along with the system number.
In step 1, there are error labels for the source and True and False labels for each edit.
In step 2, True and False labels with the system number are used to indicate whether the errors in the source were corrected.
Due to the specifications of doccano, even if the same edit appears in multiple corrections, annotators need to label each occurrence separately.
For information on Appraise in sentence-based human evaluation, you may refer to the \cite{grundkiewicz-etal-2015-human}'s work.

\section*{Acknowledgments}
We would like to express our gratitude to the action editor and anonymous reviewers for their constructive feedback.
We also thank the annotators who contributed to building the dataset.

\bibliography{tacl2021}
\bibliographystyle{acl_natbib}

\end{document}

%% file: tables/tab_example_granularity.tex
\begin{table*}[t]
\centering
\small
\begin{tabular*}{16cm}{rl}
    \Hline
    \multicolumn{2}{c}{\centerline{\textbf{Grammatical Error Correction}}} \\
    \hline
    & {\footnotesize It is hereditary.} \\
    \textbf{Source:} & Do one who suffered from this disease keep it a secret of infrom their relatives? \\
    & {\footnotesize In retrospect, its is also ones duty to ensure that he or she undergo periodic healthchecks in their own.} \\
    \hline
    \textbf{Output A:} & Should someone who suffered from this disease keep it a secret or inform their relatives? \\
    \hline
    \textbf{Output B:} & Does someone who suffers from this disease keep it a secret from their relatives? \\
    \Hline
    \multicolumn{2}{c}{\centerline{\textbf{Edit-based human evaluation}}} \\
    \hline
    \multirow{2}{*}{\textbf{Output A:}} & \bgfc{(\textbf{Rank 1})} [Do → Should] [one → someone] who suffered from this disease keep it a secret \\&[of → or] [inform → inform] their relatives? \\
    \hline
    \multirow{2}{*}{\textbf{Output B:}} & \bgfc{(\textbf{Rank 1})} [Do → Does] [one → someone] who [suffered → suffers] from this disease keep it a \\&secret [of inform → from] their relatives? \\
    \Hline
    \multicolumn{2}{c}{\centerline{\textbf{Sentence-based human evaluation}}} \\
    \hline
    \multirow{2}{*}{\textbf{Output A:}} & \bgfc{(\textbf{Rank 1})} Should someone who suffered from this disease keep it a secret or inform their \\&relatives? \\
    \hline
    \textbf{Output B:} & \bgfb{(\textbf{Rank 5})} Does someone who suffers from this disease keep it a secret from their relatives? \\
    \Hline
\end{tabular*}
\caption{
Actual data taken from our dataset shows that the results of human evaluation vary depending on the granularity.
In edit-based evaluation, output B was assigned the \bgfc{highest} rank (tied with output A), while in sentence-based evaluation, output B received the \bgfb{lowest} rank.
The results suggest that, even if all edits are considered valid, there are instances where the corrected sentence may lack fluency and naturalness in context.
}
\label{tab:granularity diff}
\end{table*}

%% file: figures/fig_m2_wer.tex
\begin{figure*}[t]
\centering
\includegraphics[width=16cm]{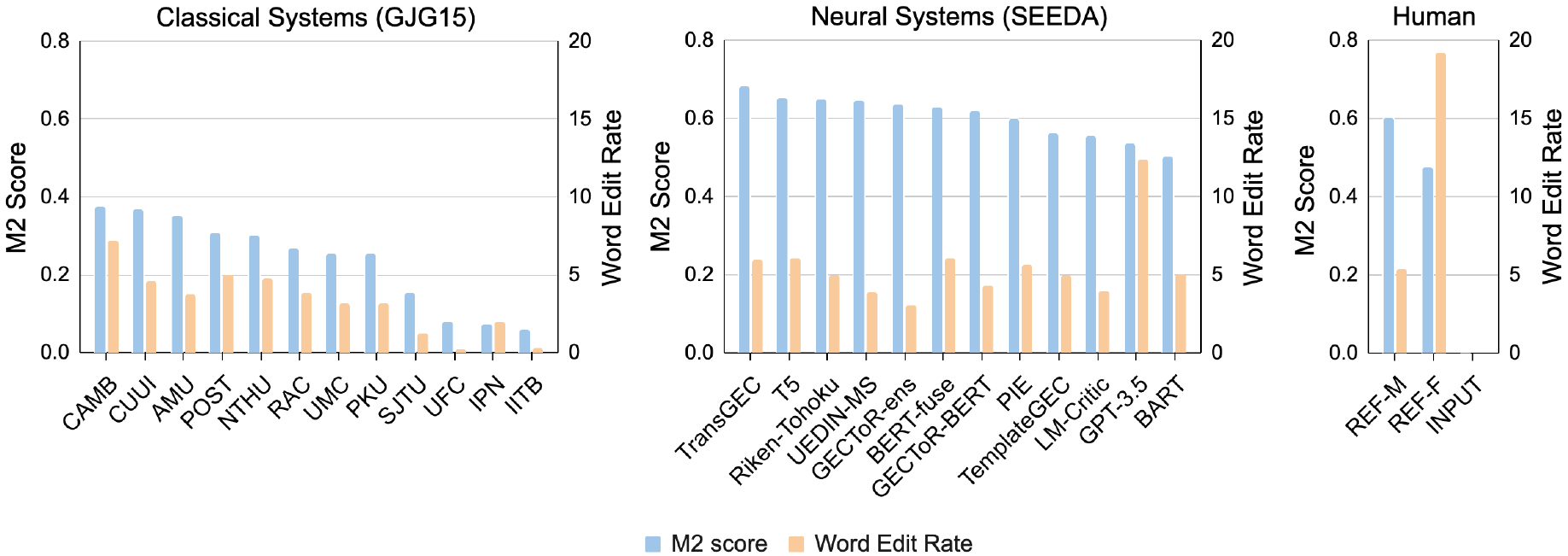}
\caption{
\mtwo Score ($F_{0.5}$) and word edit rate for classical systems in GJG15, neural systems in \seeda, and human sentences.
These neural systems generate more edits and better corrections compared to classical systems.
}
\label{fig:system analysis}
\end{figure*}

%% file: figures/fig_overview.tex
\begin{figure*}[t]
\centering
\includegraphics[width=16cm]{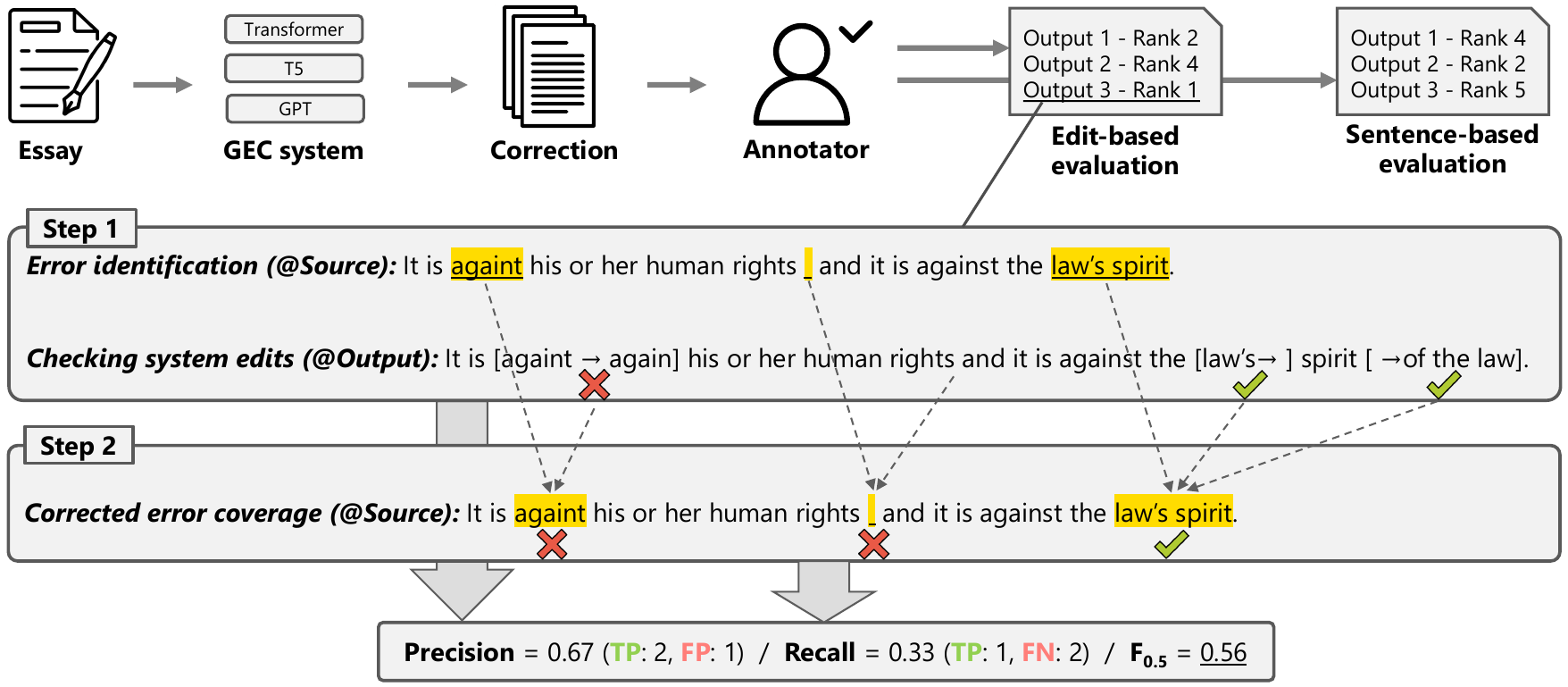}
\caption{
An overview of the annotation flow and an example of edit-based human evaluation.
In Step 1, the annotator identifies errors in the source.
Then, they categorize each edit in the output as either valid or not.
In Step 2, the annotator determines whether each edit in the output effectively corrects the errors found in Step 1.
TP, FP, and FN represent True Positive, False Positive, and False Negative, respectively.
}
\label{fig:overview}
\end{figure*}

%% file: tables/tab_human_rankings.tex
\begin{table*}[ht]
  \centering
  \scalebox{0.8}[0.8]{
  \begin{minipage}[t]{0.30\textwidth}
    \begin{center}
    \begin{tabular}{crcl}
    \Hline
    \# & Score & Range & System \\
    \hline
    1 & 0.273 & 1     & AMU \\
    \hline
    2 & 0.182 & 2     & CAMB \\
    \hline
    3 & 0.114 & 3-4   & RAC \\
      & 0.105 & 3-5   & CUUI \\
      & 0.080 & 4-5   & POST \\
    \hline
    4 & -0.001 & 6-7  & PKU \\
      & -0.022 & 6-8  & UMC \\
      & -0.041 & 7-10 & UFC \\
      & -0.055 & 8-11 & IITB \\
      & -0.062 & 8-11 & INPUT \\
      & -0.074 & 9-11 & SJTU \\
    \hline
    5 & -0.142 & 12   & NTHU \\
    \hline
    6 & -0.358 & 13   & IPN \\
    \Hline
    \end{tabular}
    \subcaption{Sentence-based evaluation in GJG15}
    \end{center}
  \end{minipage}
  \hspace{20pt}
  \begin{minipage}[t]{0.35\textwidth}
    \begin{center}
    \begin{tabular}{crcl}
    \Hline
    \# & Score & Range & System \\
    \hline
    1 & 0.992 & 1     & REF-F \\
    \hline
    2 & 0.743 & 2     & GPT-3.5 \\
    \hline
    3 & 0.179 & 3-4   & T5 \\
      & 0.175 & 3-4   & TransGEC \\
    \hline
    4 & 0.067 & 5-6   & REF-M \\
      & 0.023 & 5-7   & BERT-fuse \\
      & -0.001 & 6-8  & Riken-Tohoku \\
      & -0.034 & 7-8  & PIE \\
    \hline  
    5 & -0.163 & 9-12 & LM-Critic \\
      & -0.168 & 9-12 & TemplateGEC \\
      & -0.178 & 9-12 & GECToR-BERT \\
      & -0.179 & 9-12 & UEDIN-MS \\
    \hline
    6 & -0.234 & 13   & GECToR-ens \\
    \hline
    7 & -0.300 & 14     & BART \\
    \hline
    8 & -0.992 & 15   & INPUT \\
    \Hline
    \end{tabular}
    \subcaption{Sentence-based evaluation in SEEDA}
    \end{center}
  \end{minipage}
  \hspace{36pt}
  \begin{minipage}[t]{0.35\textwidth}
    \begin{center}
    \begin{tabular}{crcl}
    \Hline
    \# & Score & Range & System \\
    \hline
    1 & 0.679 & 1     & REF-F \\
    \hline
    2 & 0.583 & 2     & GPT-3.5 \\
    \hline
    3 & 0.173 & 3   & TransGEC \\
    \hline
    4 & 0.097 & 4-6   & T5 \\
      & 0.078 & 4-7   & REF-M \\
      & 0.067 & 4-7   & Riken-Tohoku \\
      & 0.064 & 4-7  & BERT-fuse \\
    \hline  
    5 & -0.076 & 8-11  & UEDIN-MS \\
      & -0.084 & 8-11 & PIE \\
      & -0.092 & 8-11 & GECToR-BERT \\
      & -0.097 & 8-11 & LM-Critic \\
    \hline
    6 & -0.154 & 12-12 & GECToR-ens \\
    \hline
    7 & -0.211 & 13-14   & TemplateGEC \\
      & -0.231 & 13-14   & BART \\
    \hline
    8 & -0.797 & 15   & INPUT \\
    \Hline
    \end{tabular}\\
    \smallskip
    \subcaption{Edit-based evaluation in SEEDA}
    \end{center}
  \end{minipage}
  }
  \caption{Human rankings for each evaluation granularity using TS. Systems based on GPT and T5 architectures (GPT-3.5, T5, TransGEC) consistently achieve higher rankings than REF-M, suggesting the potential for these systems to outperform human capabilities in providing corrections.}
  \label{tab:system ranking}
\end{table*}

%% file: tables/tab_stats.tex
\begin{table}[t]
\centering
\scalebox{0.8}[0.8]{
\begin{tabular}{ccc}
\Hline
Annotator & Raw data        & Expanded \\
\hline
1 & 1,777 (592 / 507) & 10,893 (6,349 / 5,919) \\
2 & 1,770 (522 / 240) & 11,663 (7,053 / 5,445) \\
3 & 1,800 (343 / 44)  & 10,988 (5,572 / 4,433) \\
\hline
Total & 5,347 (1,457 / 791)& 33,544 (18,974 / 15,797) \\
\Hline
\end{tabular}
}
\caption{
Dataset statistics for pairwise judgments by annotators.
The numbers within the parentheses represent the number of ties, with the left being edit-based and the right being sentence-based.}
\label{tab:dataset statistics}
\end{table}

%% file: tables/tab_IAA.tex
\begin{table}[t]
\centering
\small
\begin{tabular}{lll}
\Hline
Agreement & $\kappa$ (\scriptsize{\seeda /GJG15}) & Degree \\
\hline
Inter- (Edit)   & 0.28 / -  & Fair \\
Inter- (Sentence)   & 0.41 / 0.29  & Moderate \\
\hline
Intra- (Edit)   & 0.61 / -  & Substantial \\
Intra- (Sentence)   & 0.71 / 0.46 & Substantial \\
\Hline
\end{tabular}
\caption{
Cohen's $\kappa$ measures the average inter- and intra-annotator agreements on pairwise judgments.
The numbers in parentheses represent the $\kappa$ for GJG15.
}
\label{tab:annotator agreement}
\end{table}

%% file: figures/fig_scatterplots.tex
\begin{figure*}[t]
\centering
\includegraphics[width=16cm]{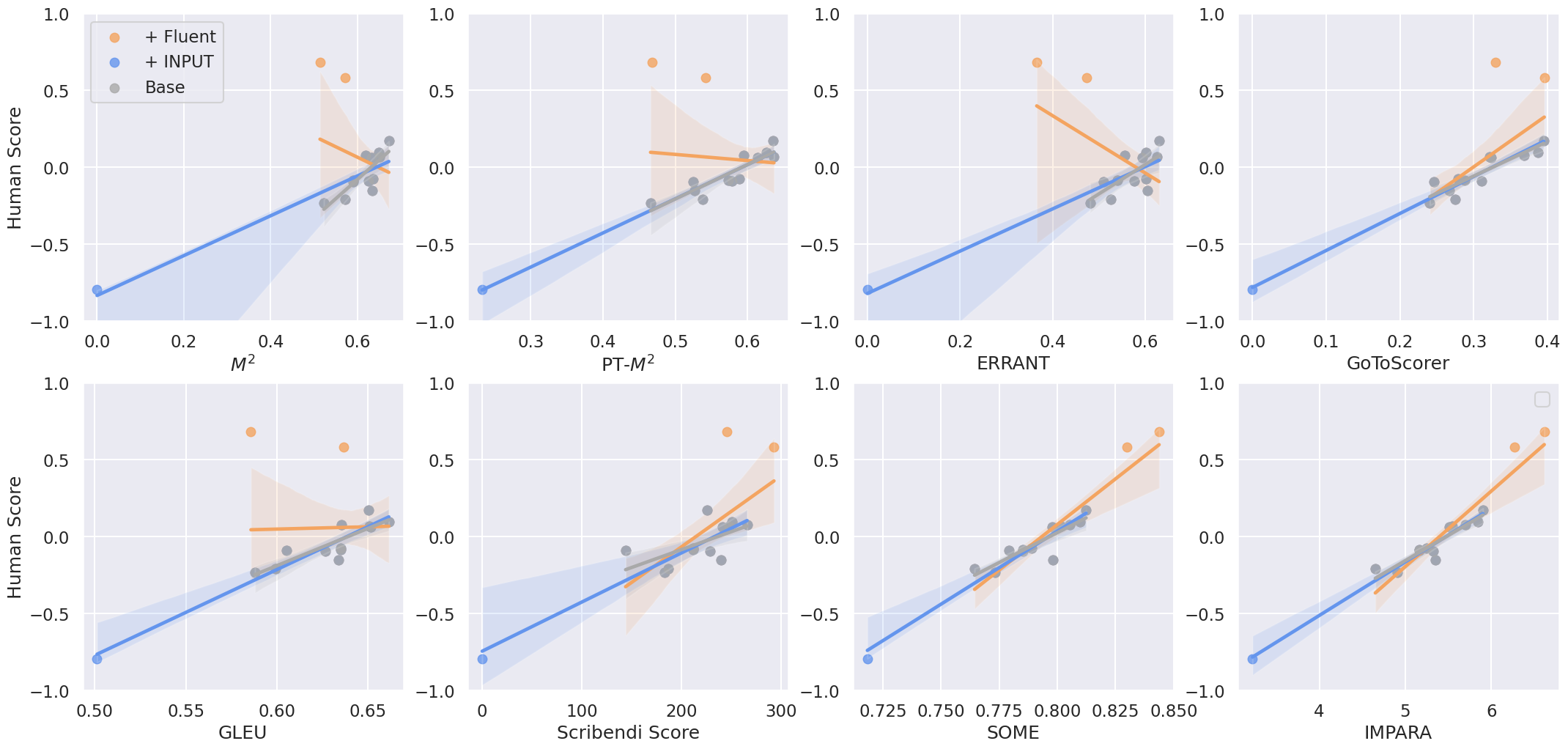}
\caption{
Scatter plots of the human score and the metric score.
``Base'' indicates the 12 systems excluding uncorrected sentences (INPUT) and fluent sentences (REF-F, GPT-3.5).
Each line represents a regression line, and the shaded area indicates the size of the confidence interval for the estimated regression, obtained using bootstrap.
Comparing the orange and blue regression lines to the gray regression line allows us to observe the degree of influence of each outlier on the distribution trend.
For example, the leftward tilt of the orange regression lines for \mtwo, \ptmtwo, ERRANT, and GLEU indicates a negative impact from fluent sentences as outliers.
}
\label{fig:scatterplot}
\end{figure*}

%% file: tables/tab_sys-level_meta-eval.tex
\begin{table*}[ht]
\centering
\scalebox{0.8}[0.8]{
\begin{tabular}{l|cc|cc|cc|cc|cc|cc}
\Hline
\multirow{3}{*}{Metric} & \multicolumn{6}{c|}{System-level} & \multicolumn{6}{c}{Sentence-level} \\
 & \multicolumn{2}{c|}{GJG15} & \multicolumn{2}{c|}{\seeda-S} & \multicolumn{2}{c|}{\seeda-E} & \multicolumn{2}{c|}{GJG15} & \multicolumn{2}{c|}{\seeda-S} & \multicolumn{2}{c}{\seeda-E} \\
 & r & $\rho$ & r & $\rho$ & r & $\rho$ & Acc & $\tau$ & Acc & $\tau$ & Acc & $\tau$ \\
\hline
 \mtwo & 0.721 & 0.706 & 0.658 & 0.487 & 0.791 & 0.764 & 0.506 & 0.350 & 0.512 & 0.200 & 0.582 & \textbf{0.328} \\
 \sentmtwo & 0.852 & 0.762 & 0.802 & 0.692 & 0.887 & 0.846 & 0.506 & 0.350 & 0.512 & 0.200 & 0.582 & \textbf{0.328} \\
 \ptmtwo & 0.912 & 0.853 & 0.845 & 0.769 & 0.896 & 0.909 & \textbf{0.512} & 0.354 & \textbf{0.527} & \textbf{0.204} & \textbf{0.587} & 0.293 \\
 ERRANT & 0.738 & 0.699 & 0.557 & 0.406 & 0.697 & 0.671 & 0.504 & \textbf{0.356} & 0.498 & 0.189 & 0.573 & 0.310 \\
 SentERRANT & 0.850 & 0.741 & 0.758 & 0.643 & 0.860 & 0.825 & 0.504 & \textbf{0.356} & 0.498 & 0.189 & 0.573 & 0.310 \\
 PT-ERRANT & \textbf{0.917} & \textbf{0.886} & 0.818 & 0.720 & 0.888 & 0.888 & 0.493 & 0.343 & 0.497 & 0.158 & 0.553 & 0.246 \\
 GoToScorer & 0.691 & 0.685 & \textbf{0.929} & \textbf{0.881} & \textbf{0.901} & \textbf{0.937} & 0.336 & 0.237 & 0.477 & -0.046 & 0.521 & 0.042 \\
\hline
 GLEU & 0.653 & 0.510 & 0.847 & \textbf{0.886} & \textbf{0.911} & 0.897 & 0.684 & 0.378 & 0.673 & 0.351 & 0.695 & 0.404 \\
 Scribendi Score & 0.890 & 0.923 & 0.631 & 0.641 & 0.830 & 0.848 & 0.498 & 0.009 & 0.354 & -0.238 & 0.377 & -0.196 \\
 SOME & \textbf{0.975} & \textbf{0.979} & 0.892 & 0.867 & 0.901 & \textbf{0.951} & \textbf{0.776} & \textbf{0.555} & \textbf{0.768} & \textbf{0.555} & \textbf{0.747} & \textbf{0.512} \\
 IMPARA & 0.961 & 0.965 & \textbf{0.911} & 0.874 & 0.889 & 0.944 & 0.744 & 0.491 & 0.761 & 0.540 & 0.742 & 0.502 \\
\Hline
\end{tabular}
}
\caption{
System-level and sentence-level meta-evaluation results excluding outliers. We use Pearson (r) and Spearman ($\rho$) for system-level and Accuracy (Acc) and Kendall ($\tau$) for sentence-level meta-evaluations.
The sentence-based human evaluation dataset is denoted \seeda-S and the edit-based one is denoted \seeda-E. 
The score in bold represents the metrics with the highest correlation at each granularity.
There is a trend of improving correlation by aligning the metrics at the sentence level (\seeda-S vs \seeda-E) and a trend of decreasing correlation by changing the target systems from classical systems to neural systems (GJG15 vs \seeda-S).
}
\label{tab:base meta-evaluation}
\end{table*}

%% file: tables/tab_effect_outlier.tex
\begin{table*}[ht]
\centering
\scalebox{0.8}{
\begin{tabular}{l|cc|cc|cc|cc|cc|cc}
\Hline
\multirow{3}{*}{Metric} & \multicolumn{6}{c|}{System-level} & \multicolumn{6}{c}{Sentence-level} \\
 & \multicolumn{2}{c|}{\small{+INPUT}} & \multicolumn{2}{c|}{\footnotesize{+REF-F, GPT-3.5}} & 
 \multicolumn{2}{c|}{\small{All systems}} & 
 \multicolumn{2}{c|}{\small{+INPUT}} & \multicolumn{2}{c|}{\footnotesize{+REF-F, GPT-3.5}} &
 \multicolumn{2}{c}{\small{All systems}} \\
 & r & $\rho$ & r & $\rho$ & r & $\rho$ & Acc & $\tau$ & Acc & $\tau$ & Acc & $\tau$ \\
\hline
    \mtwo & \textcolor[rgb]{0.1,0.7,0.3}{0.928} & \textcolor[rgb]{0.1,0.7,0.3}{0.814} & \textcolor{red}{-0.239} & \textcolor{red}{0.161} & \textcolor{red}{0.566} & \textcolor{red}{0.318} & \textcolor[rgb]{0.1,0.7,0.3}{0.605} & \textcolor[rgb]{0.1,0.7,0.3}{0.361} & \textcolor{red}{0.527} & \textcolor{red}{0.216} & \textcolor{red}{0.558} & \textcolor{red}{0.264} \\
    \mtwo (+Min) & \textcolor[rgb]{0.1,0.7,0.3}{0.929} & \textcolor[rgb]{0.1,0.7,0.3}{0.884} & \textcolor{red}{-0.172} & \textcolor{red}{0.264} & \textcolor{red}{0.587} & \textcolor{red}{0.403} & \textcolor[rgb]{0.1,0.7,0.3}{0.673} & \textbf{\textcolor[rgb]{0.1,0.7,0.3}{0.461}} & \textcolor[rgb]{0.1,0.7,0.3}{0.594} & \textcolor{red}{0.304} & \textcolor[rgb]{0.1,0.7,0.3}{0.630} & \textcolor[rgb]{0.1,0.7,0.3}{0.363} \\
    \mtwo (+Min, Flu) & \textcolor[rgb]{0.1,0.7,0.3}{0.930} & \textcolor[rgb]{0.1,0.7,0.3}{0.880} & \textcolor{red}{-0.149} & \textcolor{red}{0.262} & \textcolor{red}{0.594} & \textcolor{red}{0.400} & \textbf{\textcolor[rgb]{0.1,0.7,0.3}{0.674}} & \textcolor[rgb]{0.1,0.7,0.3}{0.458} & \textbf{\textcolor[rgb]{0.1,0.7,0.3}{0.595}} & \textbf{\textcolor{red}{0.305}} & \textbf{\textcolor[rgb]{0.1,0.7,0.3}{0.631}} & \textbf{\textcolor[rgb]{0.1,0.7,0.3}{0.364}} \\ 
    \sentmtwo & \textcolor[rgb]{0.1,0.7,0.3}{0.971} & \textcolor[rgb]{0.1,0.7,0.3}{0.879} & \textcolor{red}{-0.062} & \textcolor[rgb]{0.1,0.7,0.3}{0.358} & \textcolor{red}{0.542} & \textcolor{red}{0.479} & \textcolor[rgb]{0.1,0.7,0.3}{0.605} & \textcolor[rgb]{0.1,0.7,0.3}{0.361} & \textcolor{red}{0.527} & \textcolor{red}{0.216} & \textcolor{red}{0.558} & \textcolor{red}{0.264} \\
    \ptmtwo & \textcolor[rgb]{0.1,0.7,0.3}{0.974} & \textcolor[rgb]{0.1,0.7,0.3}{0.929} & \textcolor{red}{-0.083} & \textcolor{red}{0.442} & \textcolor{red}{0.509} & \textcolor{red}{0.546} & \textcolor[rgb]{0.1,0.7,0.3}{0.608} & \textcolor[rgb]{0.1,0.7,0.3}{0.332}  & \textcolor{red}{0.542} & \textcolor{red}{0.200} & \textcolor{red}{0.571} & \textcolor{red}{0.250} \\
    ERRANT & \textcolor[rgb]{0.1,0.7,0.3}{0.925} & \textcolor[rgb]{0.1,0.7,0.3}{0.742} & \textcolor{red}{-0.502} & \textcolor{red}{0.051} & \textcolor{red}{0.404} & \textcolor{red}{0.229} & \textcolor[rgb]{0.1,0.7,0.3}{0.597} & \textcolor[rgb]{0.1,0.7,0.3}{0.344} & \textcolor{red}{0.511} & \textcolor{red}{0.188} & \textcolor{red}{0.542} & \textcolor{red}{0.236} \\
    ERRANT (+Min) & \textcolor[rgb]{0.1,0.7,0.3}{0.922} & \textcolor[rgb]{0.1,0.7,0.3}{0.753} & \textcolor{red}{-0.462} & \textcolor{red}{0.112} & \textcolor{red}{0.475} & \textcolor{red}{0.279} & \textcolor[rgb]{0.1,0.7,0.3}{0.609} & \textcolor[rgb]{0.1,0.7,0.3}{0.350} & \textcolor{red}{0.530} & \textcolor{red}{0.184} & \textcolor{red}{0.550} & \textcolor{red}{0.218} \\
    ERRANT (+Min, Flu) & \textcolor[rgb]{0.1,0.7,0.3}{0.920} & \textcolor[rgb]{0.1,0.7,0.3}{0.725} & \textcolor{red}{-0.460} & \textcolor{red}{0.090} & \textcolor{red}{0.484} & \textcolor{red}{0.261} & \textcolor[rgb]{0.1,0.7,0.3}{0.605} & \textcolor[rgb]{0.1,0.7,0.3}{0.348} & \textcolor{red}{0.523} & \textcolor{red}{0.175} & \textcolor{red}{0.541} & \textcolor{red}{0.207} \\ 
    SentERRANT & \textcolor[rgb]{0.1,0.7,0.3}{0.965} & \textcolor[rgb]{0.1,0.7,0.3}{0.863} & \textcolor{red}{-0.357} & \textcolor{red}{0.200} & \textcolor{red}{0.354} & \textcolor{red}{0.350} & \textcolor[rgb]{0.1,0.7,0.3}{0.597} & \textcolor[rgb]{0.1,0.7,0.3}{0.344} & \textcolor{red}{0.511} & \textcolor{red}{0.188} & \textcolor{red}{0.542} & \textcolor{red}{0.236} \\
    PT-ERRANT & \textcolor[rgb]{0.1,0.7,0.3}{0.972} & \textcolor[rgb]{0.1,0.7,0.3}{0.912} & \textcolor{red}{-0.324} & \textcolor{red}{0.240} & \textcolor{red}{0.352} & \textcolor{red}{0.382} & \textcolor[rgb]{0.1,0.7,0.3}{0.580} & \textcolor[rgb]{0.1,0.7,0.3}{0.292} & \textcolor{red}{0.500} & \textcolor{red}{0.144} & \textcolor{red}{0.532} & \textcolor{red}{0.199} \\
    GoToScorer & \textbf{\textcolor[rgb]{0.1,0.7,0.3}{0.974}} & \textbf{\textcolor[rgb]{0.1,0.7,0.3}{0.951}} & \textbf{\textcolor{red}{0.667}} & \textbf{\textcolor[rgb]{0.1,0.7,0.3}{0.916}} & \textbf{\textcolor{red}{0.817}} & \textbf{\textcolor{red}{0.932}} & \textcolor{red}{0.468} & \textcolor{red}{-0.064} & \textcolor{red}{0.505} & \textcolor{red}{0.009} & \textcolor{red}{0.476} & \textcolor{red}{-0.048} \\
\hline
    GLEU & \textcolor[rgb]{0.1,0.7,0.3}{0.957} & \textcolor[rgb]{0.1,0.7,0.3}{0.911} & \textcolor{red}{-0.039} & \textcolor{red}{0.475} & \textcolor{red}{0.453} & \textcolor{red}{0.574} & \textcolor[rgb]{0.1,0.7,0.3}{0.698} & \textcolor[rgb]{0.1,0.7,0.3}{0.400} & \textcolor{red}{0.611} & \textcolor{red}{0.227} & \textcolor{red}{0.639} & \textcolor{red}{0.285} \\
    GLEU (+Min) & \textcolor[rgb]{0.1,0.7,0.3}{0.868} & \textbf{\textcolor[rgb]{0.1,0.7,0.3}{0.942}} & \textcolor{red}{0.236} & \textcolor{red}{0.704} & \textcolor{red}{0.593} & \textcolor{red}{0.760} & \textcolor[rgb]{0.1,0.7,0.3}{0.758} & \textcolor[rgb]{0.1,0.7,0.3}{0.519} & \textcolor{red}{0.662} & \textcolor{red}{0.327} & \textcolor[rgb]{0.1,0.7,0.3}{0.685} & \textcolor[rgb]{0.1,0.7,0.3}{0.372} \\
    GLEU (+Min, Flu) & \textcolor[rgb]{0.1,0.7,0.3}{0.857} & \textcolor[rgb]{0.1,0.7,0.3}{0.935} & \textcolor{red}{0.275} & \textcolor{red}{0.700} & \textcolor{red}{0.610} & \textcolor{red}{0.756} & \textcolor[rgb]{0.1,0.7,0.3}{0.756} & \textcolor[rgb]{0.1,0.7,0.3}{0.513} & \textcolor[rgb]{0.1,0.7,0.3}{0.727} & \textcolor[rgb]{0.1,0.7,0.3}{0.463} & \textcolor[rgb]{0.1,0.7,0.3}{0.684} & \textcolor[rgb]{0.1,0.7,0.3}{0.370} \\ 
    Scribendi Score & \textcolor[rgb]{0.1,0.7,0.3}{0.902} & \textcolor[rgb]{0.1,0.7,0.3}{0.718} & \textcolor{red}{0.611} & \textcolor[rgb]{0.1,0.7,0.3}{0.717} & \textcolor[rgb]{0.1,0.7,0.3}{0.755} & \textcolor[rgb]{0.1,0.7,0.3}{0.770} & \textcolor{red}{0.316} & \textcolor{red}{-0.323} & \textcolor{red}{0.345} & \textcolor{red}{-0.264} & \textcolor{red}{0.315} & \textcolor{red}{-0.328} \\
    SOME & \textcolor[rgb]{0.1,0.7,0.3}{0.965} & \textcolor[rgb]{0.1,0.7,0.3}{0.896} & \textcolor[rgb]{0.1,0.7,0.3}{0.931} & \textcolor[rgb]{0.1,0.7,0.3}{0.916} & \textbf{\textcolor[rgb]{0.1,0.7,0.3}{0.947}} & \textcolor[rgb]{0.1,0.7,0.3}{0.932} & \textbf{\textcolor[rgb]{0.1,0.7,0.3}{0.792}} & \textbf{\textcolor[rgb]{0.1,0.7,0.3}{0.601}} & \textbf{\textcolor{red}{0.760}} & \textbf{\textcolor{red}{0.531}} & \textbf{\textcolor{red}{0.766}} & \textbf{\textcolor{red}{0.537}} \\
    IMPARA & \textbf{\textcolor[rgb]{0.1,0.7,0.3}{0.975}} & \textcolor[rgb]{0.1,0.7,0.3}{0.901} & \textbf{\textcolor[rgb]{0.1,0.7,0.3}{0.932}} & \textbf{\textcolor[rgb]{0.1,0.7,0.3}{0.921}} & \textcolor[rgb]{0.1,0.7,0.3}{0.934} & \textbf{\textcolor[rgb]{0.1,0.7,0.3}{0.936}} & \textcolor[rgb]{0.1,0.7,0.3}{0.785} & \textcolor[rgb]{0.1,0.7,0.3}{0.587} & \textcolor{red}{0.742} & \textcolor{red}{0.496} & \textcolor{red}{0.745} & \textcolor{red}{0.495} \\
\Hline
\end{tabular}
}
\caption{Meta-evaluation results when an outlier is included. Green indicates an increase in correlation compared to the meta-evaluation in Table~\ref{tab:base meta-evaluation}, while red indicates a decrease.
``+Min'' in parentheses is when 11 minimal edit references are added, and ``+Flu'' is when three fluency edit references are added. 
``All systems'' is the case where all outliers are considered.
For most metrics, INPUT acts as an outlier that improves correlation, while REF-F and GPT-3.5 function as outliers that decrease correlation.}
\label{tab:outlier meta-evaluation}
\end{table*}

%% file: figures/fig_window_analysis.tex
\begin{figure*}[tbh]
\centering
\includegraphics[width=16cm]{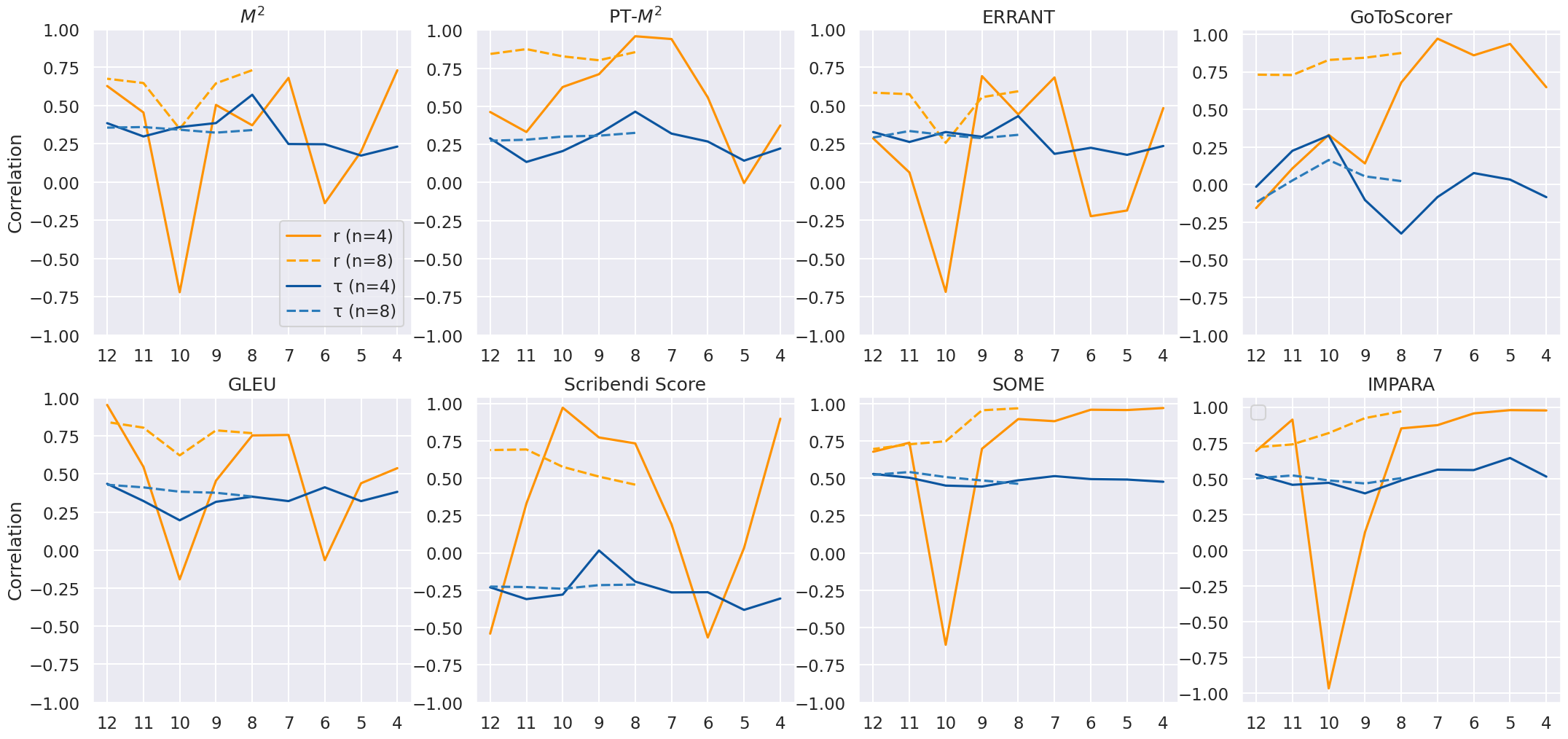}
\caption{
Variation of correlation when different systems are considered using window analysis.
The x-axis represents the human ranking of the 12 systems excluding outliers.
``n'' denotes the number of systems considered, with solid lines representing four systems and dashed lines representing eight systems. 
For example, for n$=$4, a point with x$=$5 corresponds to a human evaluation using systems ranked 2 to 5. 
The orange line represents Pearson (r) and the blue line represents Kendall ($\tau$).
The correlation of the main metrics (\mtwo, ERRANT, GLEU) shows significant variability, while pretraining-based metrics (SOME, IMPARA) exhibit relatively stable correlations.
}
\label{fig:window analysis}
\end{figure*}

%% file: figures/fig_screenshot.tex
\begin{figure*}[th]
\centering
\subfloat[Annotation Step 1.]{\includegraphics[clip, width=16cm]{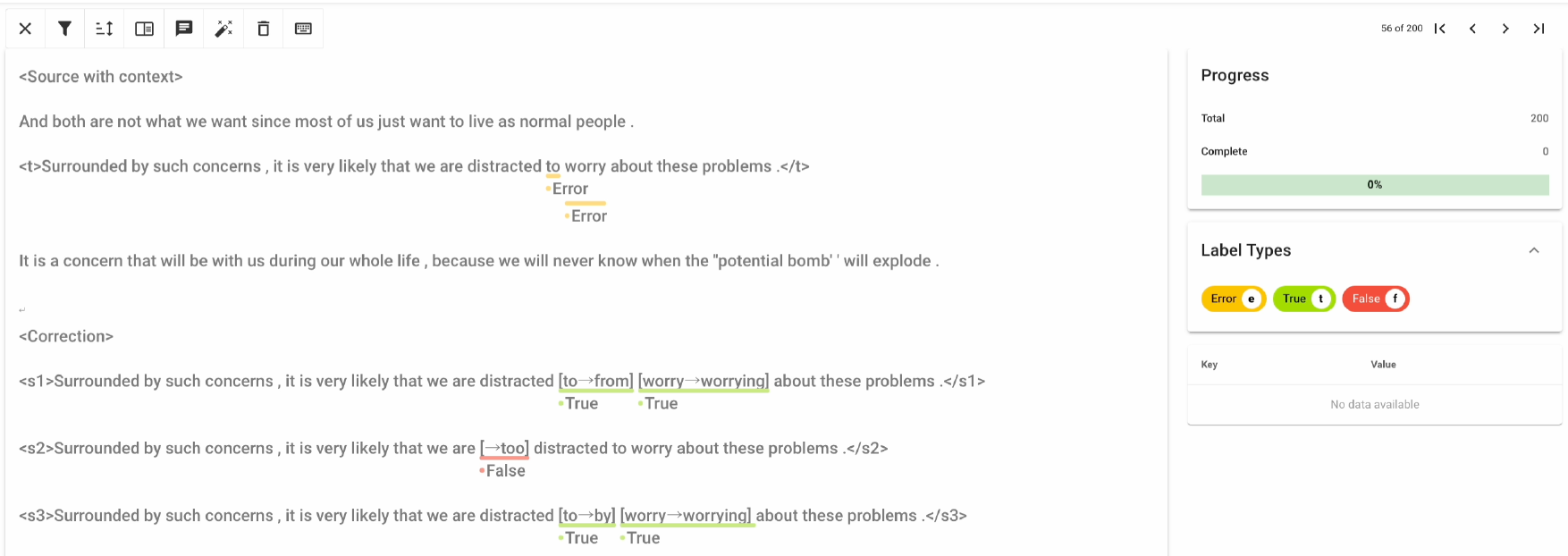}
\label{fig:step1}}\\
\subfloat[Annotation Step 2.]{\includegraphics[clip, width=16cm]{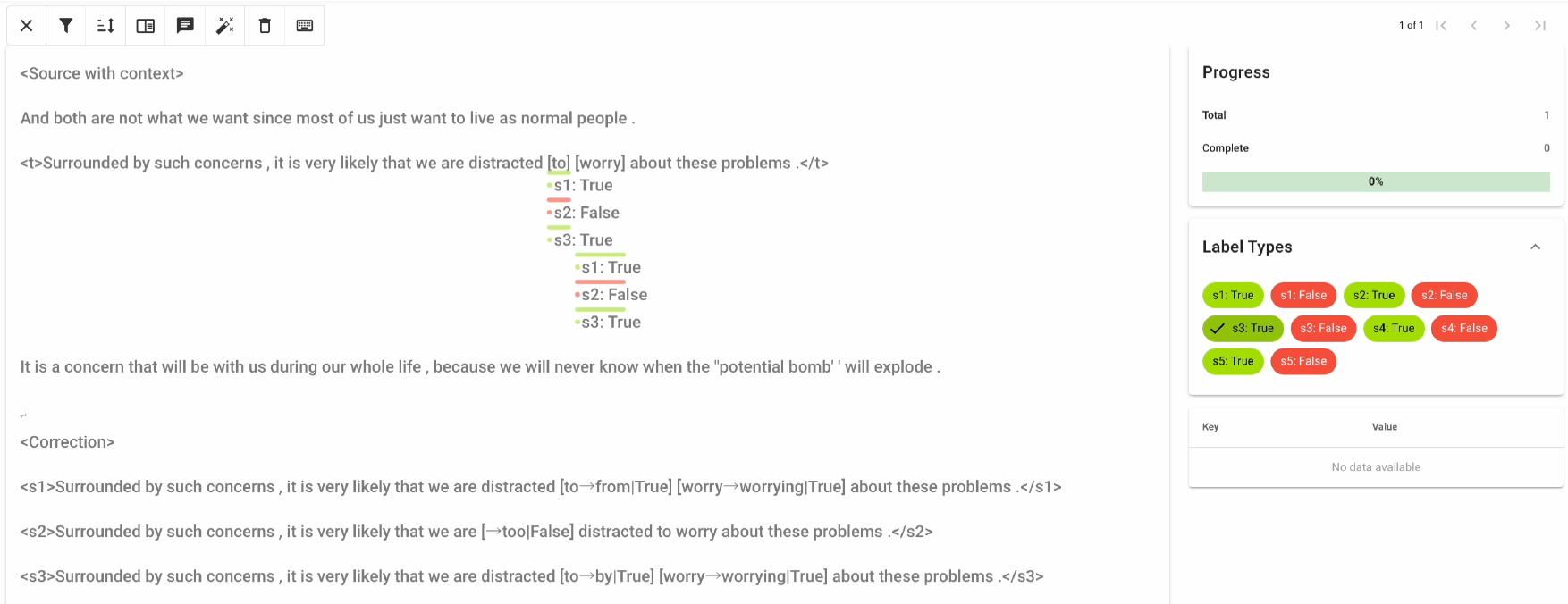}
\label{fig:step2}}
\caption{Screenshot of doccano used in the edit-based human evaluation.}
\label{fig:screenshot}
\end{figure*}